\documentclass[10pt,twocolumn,letterpaper]{article}

\usepackage{cvpr}              
\usepackage{graphicx}
\usepackage{amsmath}
\usepackage{mathrsfs}  
\usepackage{booktabs}
\usepackage{float}
\definecolor{cvprblue}{rgb}{0.21,0.49,0.74}
\usepackage[pagebackref,breaklinks,colorlinks,allcolors=cvprblue]{hyperref}

\newcommand{\xzfc}[1]{{\color[rgb]{0.0,0.0,0.0}#1}}

\newcommand{\yl}[1]{{\color[rgb]{0.0,0.0,0.0}#1}}
\def\paperID{} 
\def\confName{CVPR}
\def\confYear{2026}

\title{IM-Animation: An Implicit Motion Representation for Identity-decoupled Character Animation}

\author{Zhufeng Xu\textsuperscript{1,2}, Xuan Gao\textsuperscript{1,2}, Feng-Lin Liu\textsuperscript{1,2}, 
Haoxian Zhang\textsuperscript{3}, Zhixue Fang\textsuperscript{3},\\
Yu-Kun Lai\textsuperscript{4}, Xiaoqiang Liu\textsuperscript{3}, Pengfei Wan\textsuperscript{3}, Lin Gao\textsuperscript{1,2*}\\
\textsuperscript{1}Institute of Computing Technology, Chinese Academy of Sciences\\
\textsuperscript{2}University of Chinese Academy of Sciences\\ 
\textsuperscript{3}Kling Team, Kuaishou Technology
\textsuperscript{4}Cardiff University\\
}

\begin{document}

\twocolumn[{%
\renewcommand\twocolumn[1][]{#1}%
\maketitle
\begin{center}
    \centering
    \captionsetup{type=figure}
    \includegraphics[width=1\textwidth]{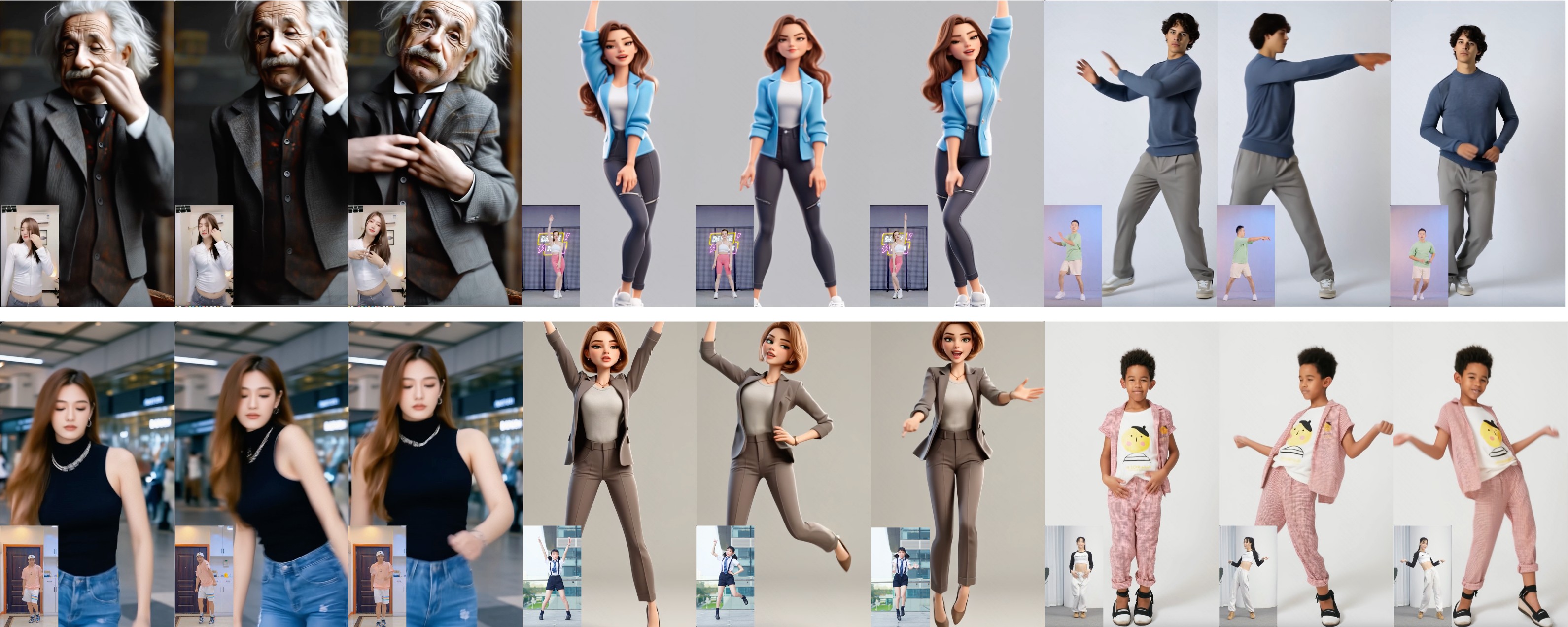}
    \captionof{figure}
    {IM-Animation introduces an impressive implicit motion representation and retargeting method. Our model supports implicit video model motion control in cases with significant 
    scale
    differences or substantial variations in posture and body \xzfc{shape}.
    \label{fig:teaser}
}
\end{center}%
}]
\begin{abstract}
{Recent progress in video diffusion models has markedly advanced character animation, which synthesizes motioned videos by animating a static identity image according to a driving video. 
Explicit methods represent motion using skeleton, DWPose or other explicit structured signals, but struggle to handle spatial mismatches and varying body \yl{scales}. 
Implicit methods, on the other hand, capture high-level implicit motion semantics directly from the driving video, but suffer from identity leakage and entanglement between motion and appearance.}
To address {the above} challenges, we propose a novel implicit motion representation {that compresses per-frame motion into compact 1D motion tokens. This design relaxes strict spatial constraints inherent in 2D representations and effectively prevents identity information leakage from the motion video. Furthermore, we} 
design a temporally consistent mask token-based retargeting module {that enforces a temporal training bottleneck, mitigating interference from the source image’s motion and improving retargeting consistency.}
Our methodology employs a three-stage {training} strategy {to enhance the training efficiency and ensure high fidelity. Extensive experiments demonstrate}
that our {implicit} motion representation and {the proposed} IM-Animation 
{achieve superior or} competitive {performance compared} with state-of-the-art methods.
\end{abstract}

\section{Introduction}
\label{sec:intro}

Character image animation is a popular research topic with \yl{a wide range of} applications in gaming, film production, and virtual reality.
This task aims to precisely transfer motion and expressions from a driving video to a new identity provided by a source image.
Early works~\cite{siarohin2019first,wang2019few} built upon GAN-based \yl{frameworks}~\cite{goodfellow2014generative}, but suffered from low resolution, poor generalization, and unrealistic, fuzzy details.

Recently, diffusion models \yl{have} emerged as a powerful solution for video generation tasks~\cite{singer2022make,guo2023animatediff,ho2022video,chen2023videocrafter1,chen2024videocrafter2,blattmann2023stable}, including character animation.
Large video foundational models ~\cite{bar2024lumiere,blattmann2023stable,blattmann2023align,brooks2022generating,guo2023animatediff,gupta2024photorealistic,ho2022video,li2018video,lin2025diffusion,singer2022make,villegas2022phenaki,wang2023modelscope,wang2020imaginator,zhou2022magicvideo,zhou2024allegro} have achieved remarkable progress in text- or image-to-video synthesis, serving as versatile backbones for downstream tasks.
However, a major challenge remains: how to precisely control content and style to satisfy personalized requirements across varied contexts.
Recent efforts~\cite{guo2024sparsectrl, lin2024ctrl} integrate control signals via modules such as  ControlNet~\cite{zhang2023adding}, ControlNet++~\cite{li2024controlnet++}, and T2IAdapter~\cite{mou2024t2i}.
With the development of Diffusion \yl{Transformer} (DiT)~\cite{peebles2023scalable}, how to effectively inject condition signals while avoiding cumbersome network becomes an interesting problem. 

For video diffusion-based character animation, {besides control sign injection}, another core challenge lies in effectively decoupling the identity and motion information.
Ideally, the target character’s body shape, spatial layout, and appearance should remain unaffected by those in the driving video.
Similarly, the final retargeting pose should not be influenced by the motion in the source image. 
This issue is mild in self-reenactment tasks~\cite{zuo2025dreamvvt,trymagictryon,zheng2024dynamic,chong2025catv2ton}, such as virtual try-on, because the characters in the driving video and edited image have similar body shape and \yl{scale}. 
In contrast, cross-reenactment—e.g., animating a tall adult from a child’s motion, or vice versa—poses substantial difficulty due to drastic mismatches in body shape and spatial layout.

Existing cross reenactment works can be classified into explicit and implicit methods.
Explicit approaches~\cite{hu2024animate,zhao2025dynamictrl,wang2025unianimate,luo2025dreamactor,tan2024animate,zhu2024champ,wang2024vividpose,zhou2024realisdance,zhou2025realisdance} use representation such as skeleton, DWPose~\cite{yang2023effective} or SMPL~\cite{SMPL:2015,SMPL-X:2019} to capture motion signals, but cannot handle the mentioned identity conflict between the driving video and source image. 
Follow\yl{-up} work~\cite{wang2024unianimate} \yl{rescales} the joint \yl{lengths} of the source identity to match the driving person, but \yl{fails} in complex movements.
Other approach~\cite{cheng2025wan} utilizes \yl{an} image model to change \yl{the} source image into a standard pose for shape alignment, while introducing additional cumulative errors.
Implicit works go beyond learning basic dynamic patterns to capture deeper and more semantic understandings of motion rather than merely replicating spatial shifts.
However, most of them~\cite{wei2024dreamvideo,li2024motrans,ma2025follow,zhang2025flexiact,tan2025synmotion} rely on per-video fine-tuning, making them time-consuming and restricting practical application scope.
More recent generalized method~\cite{song2025x} directly \yl{uses} 2D character tokens to represent identity, but \yl{carries} the risk of leaking motion information that \yl{affects} the retargeting pose accuracy.

We propose IM-Animation, a diffusion-based framework that achieves robust character animation even when the source and driving identities differ significantly in body shape or spatial layout (e.g., full-body driving half-body), as illustrated in Figure~\ref{fig:teaser}.
To extract motion while preventing identity leakage, we design a compact motion representation that compresses per-frame dynamics into 1D motion tokens via a transformer-based encoder–decoder with a quantized codebook.
This spatially invariant representation, trained with 
keypoint supervision, relaxes 2D grid constraints to remove identity information in the driving video.
To merge the driving motion with the source identity while preventing motion leakage, we design a novel mask token based retargeting module.
By incorporating learnable mask tokens as a training bottleneck within self-attention, our module removes 
pose information from the source 
{image to ensure} coherent motion transfer and pose consistency.
The resulting retargeted latent features and expression cues are then fed into a video diffusion model to generate realistic animation sequences.
Finally, a three-stage training strategy progressively optimizes motion representation, retargeting module, and video generation, leading to stable training and high-quality outputs.

Extensive experiments show that IM-Animation achieves high-quality results with limited computational resources. Our main contributions are:
\begin{itemize}[leftmargin=*]
\item 
We present an innovative compact motion representation that compresses each frame's motion into spatially invariant 1D tokens, effectively preventing character information leakage while preserving animation integrity.
\item
We introduce a mask token based retargeting module that leverages mask tokens as a latent bottleneck to remove motion information from the source image, ensuring retargeting pose coherence and consistency.
\item
We develop a three-stage training pipeline that progressively trains motion representation, retargeting module, and video diffusion model. 
It reduces training costs, achieves effective disentanglement and produces realistic animation results.
\end{itemize}

\section{Related Work}
\label{sec:related_work}

\subsection{Diffusion Model for Video Generation}

\xzfc{Recently, diffusion models have become a key player in visual generation, transforming noise into high-fidelity content through iterative denoising. Research has increasingly focused on enhancing their efficiency, leading to milestones like Latent Diffusion Models (LDMs)~\cite{rombach2022high}. In this evolving field, video generation has emerged as a significant challenge, emphasizing temporal coherence and dynamic scene modeling.}
Early explorations {\yl{extended} text-to-image (T2I) models to video generation.}
{Some} studies~\cite{esser2023structure,ho2022video,hong2022cogvideo,khachatryan2023text2video,qi2023fatezero,singer2022make,wu2023tune,yang2023rerender} 
{applied} inter-frame attention mechanisms, enabling models to capture dynamic dependencies.
Other research introduced dedicated temporal layers~\cite{blattmann2023align} or developed injectable motion modules~\cite{guo2023animatediff}. These modules were trained on large video datasets to integrate temporal dynamics into T2I backbones without sacrificing pre-learned spatial knowledge.

{The video generation paradigm significantly changes} with the rise of Diffusion Transformers (DiTs)~\cite{peebles2023scalable} as a mainstream architecture. 
Unlike previous structure, 
DiTs leverage self-attention mechanisms to model global spatio-temporal dependencies. This allows them to capture long-range temporal coherence and fine-grained spatial details. Building on this, recent state-of-the-art works~\cite{zheng2024open,yang2024cogvideox,kong2024hunyuanvideo,wan2025wan} have further advanced DiT-based video generation, achieving remarkable results in high-fidelity content creation and temporal consistency by scaling model capacity and training on massive text-video datasets.

\subsection{Video-driven Character Animation}

\xzfc{Video models excel at generating detail-rich results and serve as a foundation for complex controllable tasks. In character animation, they are widely used to create high-quality dynamic representations. 
}
Based on the form of control signal representation, existing methods can generally be categorized into two main types: explicit control signal  and implicit control signal \yl{representations}.

\subsubsection{Explicit Motion Representation}
{Explicit methods extract} pose sequences from the driving video, which serve as explicit guides for animating the target character. 
For instance, numerous works leverage skeletal keypoints as conditional signals~\cite{hu2024animate,zhao2025dynamictrl,wang2024unianimate,wang2025unianimate,luo2025dreamactor,tan2024animate}, while others 
{adopt} more detailed representations such as DensePose maps~\cite{xu2024magicanimate}, SMPL renderings~\cite{zhu2024champ,wang2024vividpose,zhou2024realisdance,zhou2025realisdance}, or specially designed motion cues~\cite{xu2024high,luo2025dreamactor,gao2025conmo,geng2025motion,gu2025diffusion} to better capture desired movements.

However, a critical limitation of these methods lies in their sensitivity to spatial discrepancies: when there exists a significant shape gap between the target character and the subject of the driving video, the generated animations often suffer from distortions like misaligned limbs or unnatural postures.
To mitigate this issue, researchers have proposed 
{some interesting} solutions. 
UniAnimate-DiT~\cite{wang2025unianimate} and RealisDance-DiT~\cite{zhou2025realisdance} introduce pose alignment as a preprocessing step to align the driving motion with the body of the target character, while Animate-X~\cite{tan2024animate} learns an additional implicit pose indicator by merging CLIP features and DWPose keypoints from the driving video.
However, these methods still struggle to handle complex motion scenarios, which can result in failures in the driving process.

\subsubsection{Implicit Motion Representation}

{Instead of strict spatial constraints, another line of works learn implicit motion representations directly from driving videos. 
They enable flexible cross-subject motion transfer by capturing dynamic patterns and temporal correlations in a data-driven manner.
Firstly, this strategy succeeded in facial animation works~\cite{xu2024vasa,drobyshev2024emoportraits,ki2024float,xie2024x,zhao2025x,xu2025hunyuanportrait}, which}
excel at capturing subtle expression changes in the latent space. 
{However}, extending implicit motion representation to full-body digital humans poses greater challenges{, as it demands coherent modeling of complex, multi-part body movements and fine-grained details across both space and time.}

{To solve the above issue, two categories of} approaches have emerged: finetuning-based methods and training-free methods.
Finetuning-based approaches~\cite{wei2024dreamvideo,li2024motrans,ma2025follow,zhang2025flexiact,tan2025synmotion} adapt the model to each specific driving video through fine-tuning, encoding motion information into the model weights to guide animation. This per-video 
{fine-tuning} limits efficiency for full-body scenarios.
In contrast, training-free methods~\cite{meral2024motionflow,chen2025lmp} avoid the 
{costly} per-video tuning by designing attention-based operations to directly extract motion features from the driving video during inference, but often struggle with preserving motion precision.  

\xzfc{To address these limitations, recent works have explored end-to-end frameworks to enhance efficiency and transferability for full-body applications. EfficientMT~\cite{cai2025efficientmt} reuses a pretrained video model for reference feature extraction, enabling end-to-end motion transfer. However, it lacks support for driving specific characters and struggles with complex motion \yl{sequences}. X-UniMotion~\cite{song2025x} uses dedicated image encoders to extract motion tokens for the full body, hands, and face separately, retargeting them to the reference identity via a ViT decoder for fine-grained, identity-preserving motion transfer. However, it directly models \yl{patchified} token sequence of \yl{the} character image, which can lead to shortcut learning in motion representations.}


\section{Methods}
\begin{figure*}[h!]
    \centering
    \includegraphics[width=1\textwidth]{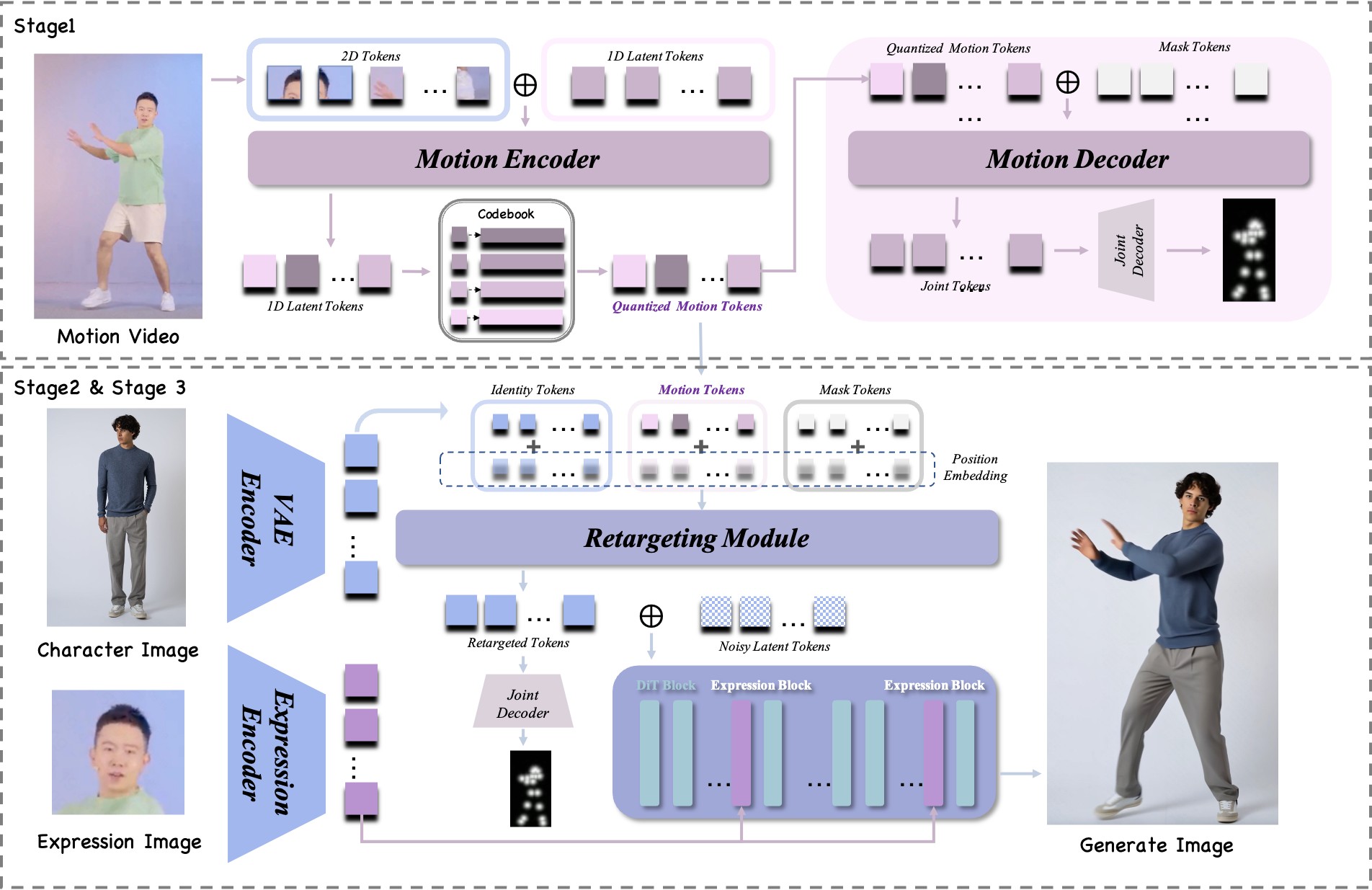} 
    \caption{We propose IM-Animation, an implicit portrait animation solution. Given a identity image and a motion video, we employ a three-stage training strategy. In the first stage, we train a compact motion encoder based on \yl{a} 1D tokenizer. In the subsequent second and third stages, we train a temporal retargeting module based on mask tokens, utilizing a lightweight heatmap decoder for intermediate supervision. This approach ensures that we can encode precise retargeted information without disclosing the ID information of the driven video or the pose information of the source image. Ultimately, we achieve end-to-end training of the entire model.}
    \label{fig:method}
    \vspace{-1em}
\end{figure*}

\begin{figure}[h!]
    \centering
    \includegraphics[width=0.45\textwidth]{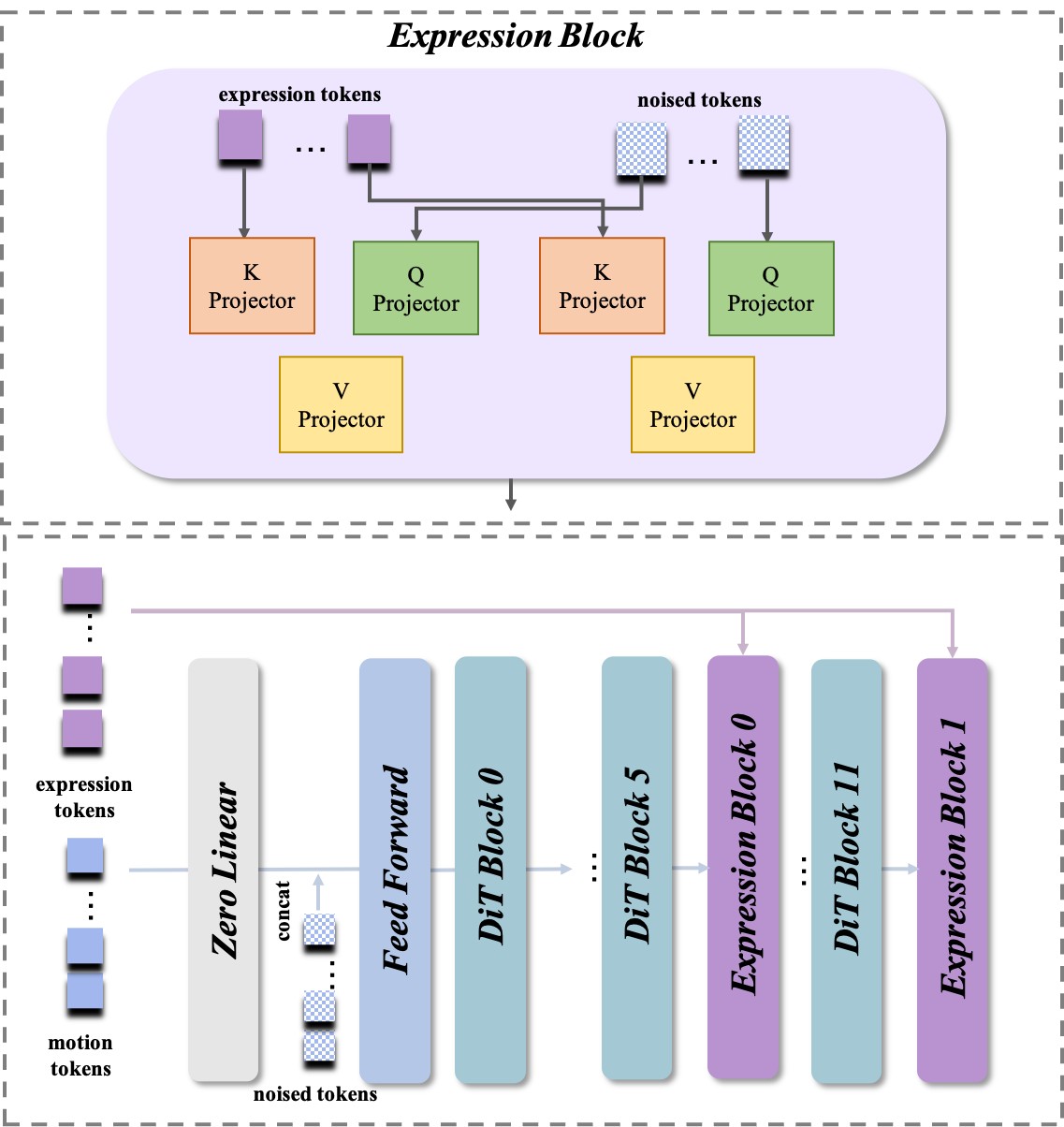} 
    \caption{Method of Control Signal Injection in Video Model.}
    \label{fig:control}
    \vspace{-1em}
\end{figure}

Given a identity image and a motion-driven video of any character, 
{our method generates a retargeted video} that matches the body shape and perspective of the character identity{ following the given driving motion}. 
{To achieve effective decoupling, we design a novel motion representation extracted from} \yl{a} video clip while preventing the leakage of identity information 
{from} the motion sequence. 
\xzfc{Furthermore, to integrate identity information into the motion feature}, we retarget the motion representation to the given character image, which provides information about the character's identity, pose and spatial location. 
\xzfc{We establish a learning bottleneck for the model by injecting a mask token to prevent it from learning the original motion representations of the identity image, which could lead to the leakage of motion information contained in the identity image.}
\xzfc{Then we use a video model to generate the video based on the compact retargeted identity representation.}

\subsection{Motion Representation}
\label{subsec:Motion Representation}
\xzfc{We do not require the motion video to provide detailed spatial correspondences. In other words, we hope that the motion tokens can transcend the original 2D grid areas after patchification, achieving more flexible, high-level, and semantically rich compact representations that are not limited to 2D space.}
Inspired by the work of TiTok~\cite{yu2024an}, which has achieved remarkable results in image reconstruction and generation, we design 
{a} motion representation encoder to resemble \yl{a} 1D tokenizer
\xzfc{which can obtain a more compact one-dimensional representation of tokens that is independent of patches}.
This design enables the encoding of each frame of the original driving video into a compact and high-quality motion representation comprising tokens.

Specifically, we 
{concatenate a set of learnable tokens of length $N_{\text{m}}$ with the patchified motion video frames, \yl{and} then feed them} into a vision transformer (ViT) encoder. In our settings, $N_{\text{m}} = 32$. 
Subsequently, the corresponding components of the learnable tokens output by motion encoder $\mathscr{E}_{\rm motion}$ are sent to a codebook for vectorization, resulting in the formation of 1D tokens, which can be defined as:
\begin{equation}
    M_{\rm img} = Quant({\mathscr{E}_{\rm motion}(T_{\rm img}\oplus T_{\rm latent})})\in \mathbb{R}^{N_{\text{m}}\times C_{\text{m}} }
\end{equation}
Here, $M_{\rm img}$ represents the final encoded motion representation, with a length of $N_{\text{m}}$ and  dimension of $C_{\text{m}}$. $T_{\rm img}$ and $T_{\rm latent}$ refer to patchified image tokens and learnable latent tokens, respectively. $Quant$ is the quantizer with a codebook that vectorizes the output of the motion encoder. The following operations are performed during this process:
\begin{equation}
Quant(z_{\rm 1d}) = c_i, \ \text{where} \ i = \operatorname*{argmin}_{j \in \{1, 2, \ldots, K\}} \|z_{\rm 1d} - c_j\|_2.
\end{equation}
\begin{figure*}[h!]
    \centering
    \includegraphics[width=1\textwidth]{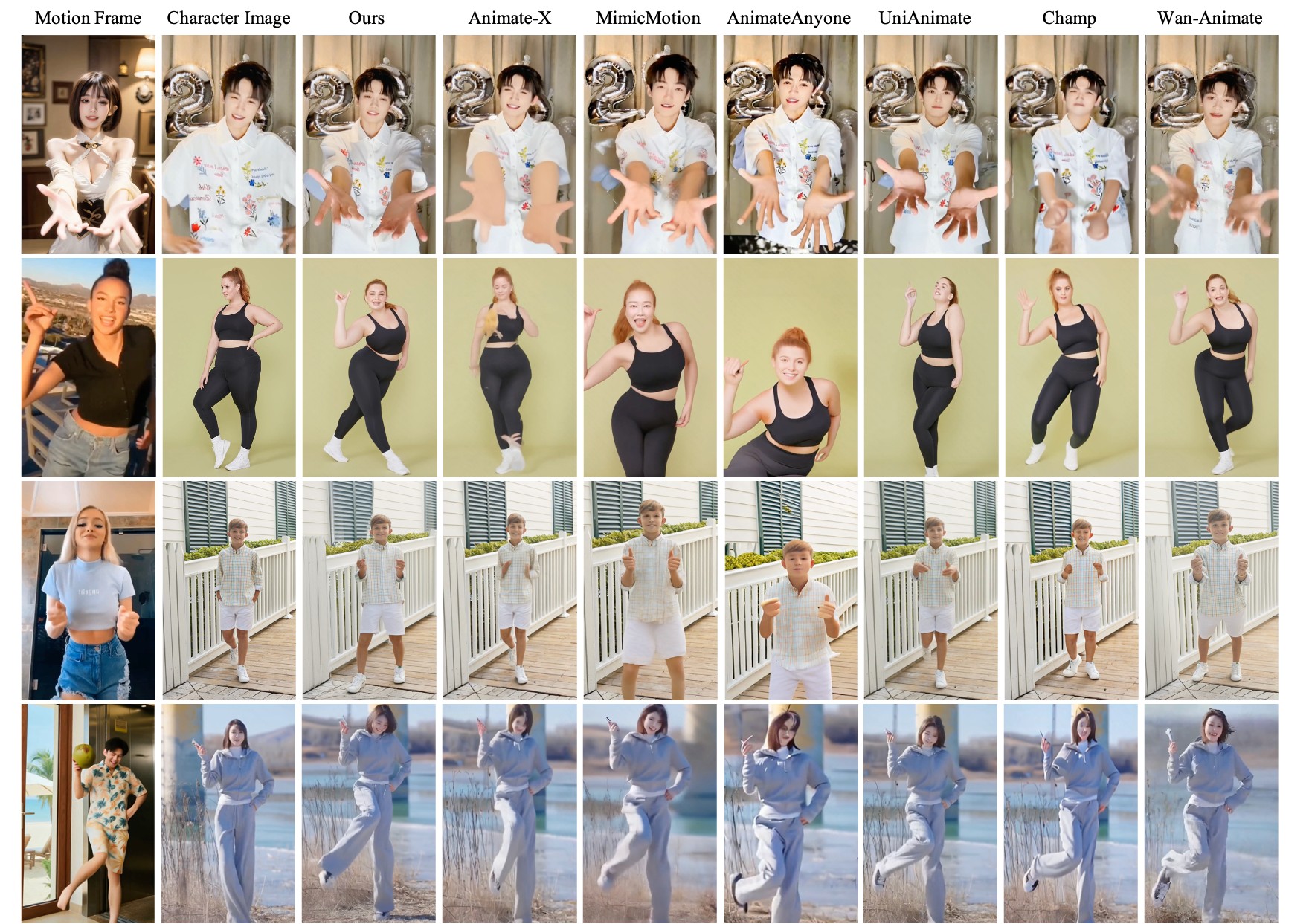}

    \caption{\textbf{Qualitative Results.} We compare our method with several state-of-the-art approaches, among which UniAnimate-DiT and Wan-Animate are trained using a 14B base model. In contrast, our method is faster while also achieving competitive results. Furthermore, our approach demonstrates impressive performance in maintaining character identity and retargeting significant character differences.}%
    \label{fig:exp1}
    \vspace{-1em}
\end{figure*}%
Here, $z_{\rm 1d}$ denotes the retained latent tokens output by the Transformer encoder, which are generated based on two sets of image tokens and learnable tokens during this process. The vector quantizer $Quant$ \yl{maps} it to the nearest code $c_i \in \mathbb{R}^{C_{\rm m}}$ in the learnable codebook \(\mathbb{C} \in \mathbb{R}^{K \times C_{\rm m}}\).

Unlike 
{reconstruction or generation} tasks, 
the motion encoder {should not retain any} 
character 
{shape or} appearance texture features 
{apart} from motion.
\xzfc{We model the first training stage as a  task which regresses the joint map from the motion video.}
{ We carefully design a training process with the help of an additional 
\xzfc{motion} decoder, which \yl{is} only used in training Stage 1
\yl{rather than inference.}
We repeat the mask token 
{$J$ times} corresponding to the number of joints, denoted as $T_{\rm mask} \in \mathbb{R}^{J \times D}$. 

Similar to the design of the {motion} encoder
, in order to supervise the extraction of motion features in the first training stage, \( M_{\text{img}} \) and \( T_{\text{mask}} \) 
{are} concatenated and fed into the transformer decoder. 
\xzfc{In Stage 1, the parts of the output other than \( T_{\text{mask}} \) are dropped, and a joint decoder \( \mathscr{D}_{\text{joint}} \) are applied to adjust the output to match the size of the corresponding mid-level supervised joint heatmap. This process is supervised by the Mean Squared Error (MSE) loss between the output joint heatmap and the real joint heatmap.
}

\begin{equation}
{\hat {H}_{\rm joint}} = \mathscr{D}_{\rm joint}(\mathscr{D}_{\rm motion}(M_{\rm img} \oplus T_{\text{mask}})).
\end{equation}
\subsection{Temporal Retargeting}


\yl{We} hope that the motion representation extracted by the motion encoder can provide feature information, while the character image offers identity and spatial layout information. 
We design a retargeting module based on mask tokens.
To prevent the character image from leaking motion information, we set learnable mask tokens, which are concatenated with image latent tokens and temporal motion tokens before being fed into the retargeting module for training.

In order to effectively distinguish between the three different types of input tokens, we additionally apply a set of learnable  positional embeddings for each type of tokens,
\begin{equation}
    M_{\rm ret} = {Retarget}(M_{\rm video}\oplus T_{\rm ref} \oplus T_{\rm mask})\in \mathbb{R}^{\frac{W}{f}\times \frac{H}{f}\times D }
\end{equation}

Here, \( M_{\text{video}} \) is the motion encoding representation of the entire sequence obtained by concatenating \( M_{\text{img}} \) along the temporal dimension. \( T_{\text{mask}} \)  \yl{is} the learnable tokens aligned in dimension and  \( T_{\text{ref}} \) \yl{is} the features of the reference image that have been encoded through VAE and 
\yl{patchification},
respectively. In Stage 2, we use $\mathscr{D}_{joint}$ to train both motion representation and retargeting module at the same time.
\xzfc{\subsection{Control Signal Injection}
We believe that incorporating the video model as a flow matching decoder can obtain more advanced semantic features during the training phase. Therefore, in the end-to-end process, we ultimately release all the aforementioned trainable parameters for further fine-tuning.

After obtaining the retargeted motion tokens, we stack them with the noisy latent representations along the channel dimension and perform patch embedding to ensure their size is supported by the video model. For facial expressions, we aim for our method to transfer the expression sequences from the original video. We use the  encoder from X-NeMo~\cite{zhao2025x} as our expression encoder and calculate cross attention frame by frame to inject the expression signals. The final output is then connected via a residual connection to the corresponding DiT Block output.
}
\subsection{Training Strategy}
{Most of existing} implicit video-driven methods require substantial training costs. 
{In contrast, w}e aim to achieve more efficient convergence under the constraint of limited training resources.

Overall, we train IM-Animation in three stages. Inspired by X-Unimotion~\cite{song2025x}, we implement mid-level joint heatmap supervision to ensure that the model can learn effective motion representations.
\begin{itemize}[leftmargin=*]
    \item \textbf{Stage 1.} We first train the motion encoder, as introduced in \yl{Sec.~}\ref{subsec:Motion Representation}. The joint-aligned decoder is only trained in this phase. During this stage,  we fine-tune the pretrained encoder model and quantizer from TiTok  using the self-reconstructed joint map for supervision.
    
    \item \textbf{Stage 2.} During the joint training of the motion encoder and the retargeting module, we \yl{perform} color augmentation, random cropping, and padding on the video data collected from the internet. Additionally, we \yl{construct} a batch of action-consistent but identity-inconsistent paired datasets using Unreal Engine (UE). To ensure that the retargeting module can  predict the actions of the reference image identity, 
    We also apply \yl{the} joint decoder here to decode joint map for supervision.

    \item \textbf{Stage 3.} 
    Finally, we inject the motion and expression control signals into the DiT model to achieve end-to-end training in the final stage.
    During this process, we also \yl{maintain} the supervision from the joint decoder.
    
\end{itemize}

\section{Experiments}
\subsection{Implementation Details}
\textbf{Datasets.} We collected over 50k video clips for our training set. To enable the model to effectively retarget characters of varying body types, we generated and selected 8k data pairs with identical motions but different identities using Unreal Engine 5 and other generation models.
In the comparative experiments, we use 50 samples from the data we collected, which {excludes in the training set, to evaluate three key aspects: cross-identity driving accuracy, generated video quality, and motion precision.}
To further demonstrate the robustness of our method, we test the performance of the Self-Reenactment task on a publicly available TikTok dataset~\cite{Jafarian_2021_CVPR_TikTok,Jafarian_2022_TPAMI} and compare with existing work.
\begin{figure}[t!]
    \centering
    \includegraphics[width=0.45\textwidth]{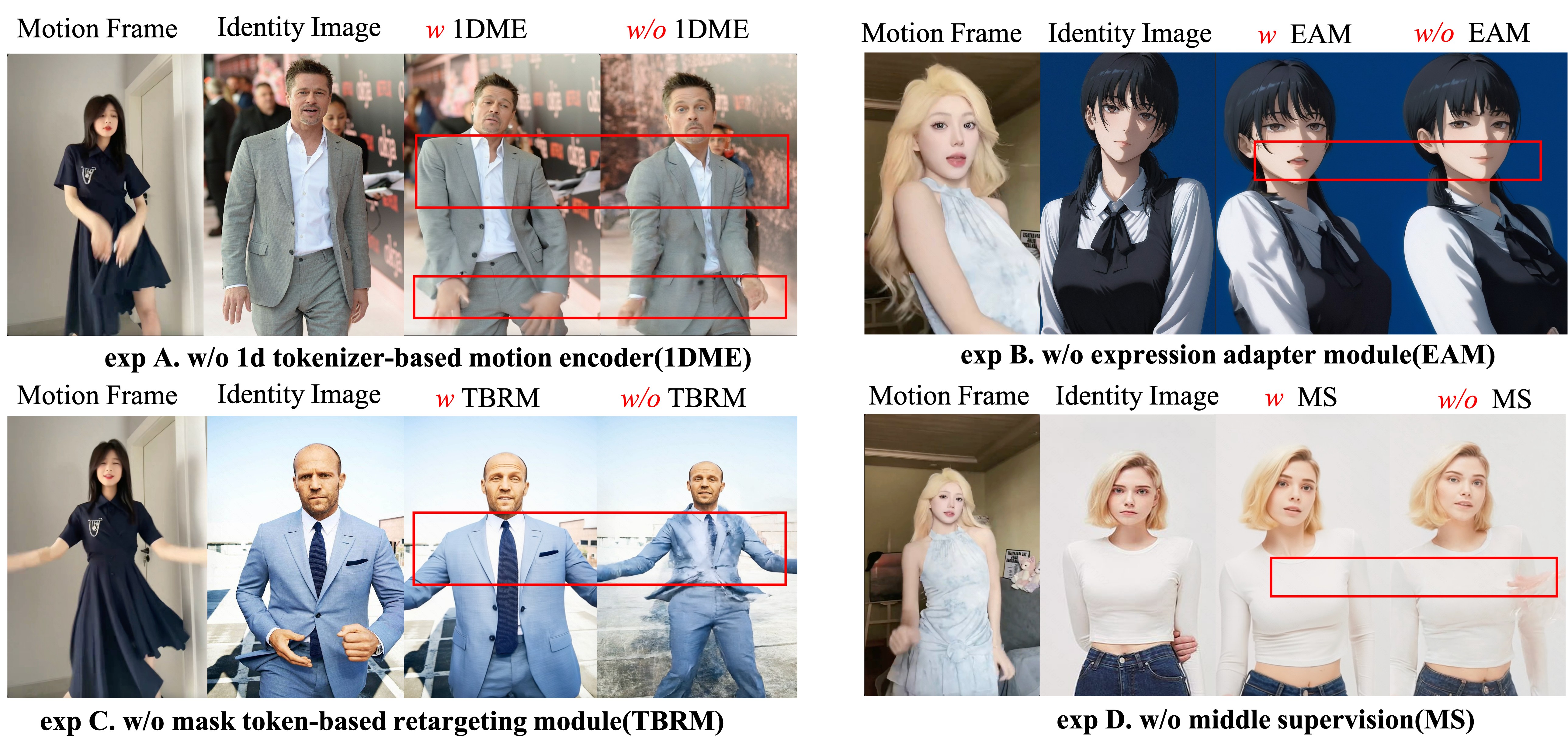} 
    \caption{Qualitative Results of Ablation Experiment. In each set, the images are arranged from left to right as follows: driving frame, character image, full model performance, and performance of the ablation experiment without the specified module.}
    \label{fig:abl}
    \vspace{-1em}
\end{figure}
\begin{figure*}[t!]
    \centering
    \includegraphics[width=1\textwidth]{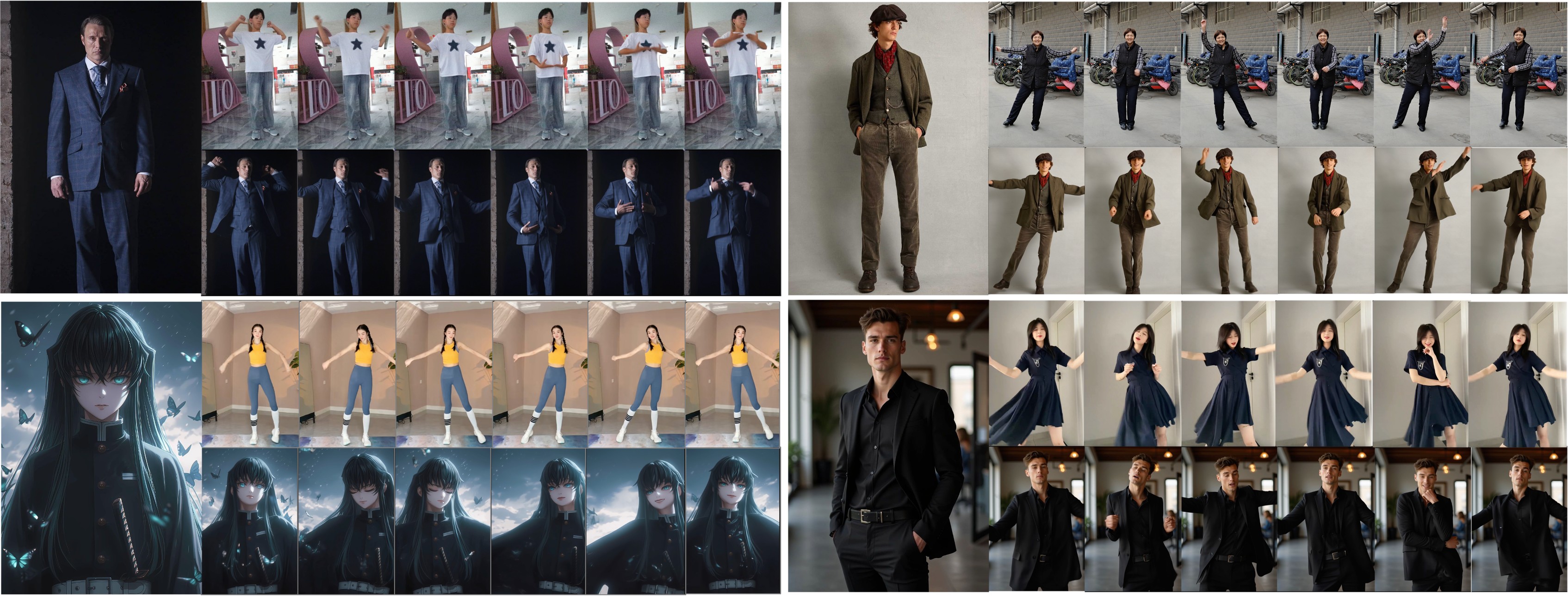} 
    \caption{Qualitative Results of IM-Animation. The top row of each image set represents the driving video, while the bottom row displays the generated results.}
    \label{fig:in the wild performance}
    \vspace{-1em}
\end{figure*}

\textbf{Training Settings.}
All of our experiments are conducted in an environment with 16 GPUs.
For the motion encoder in Stage 1, we train the self-reconstruction model for 5k
steps. In Stage 2, we train the retargeting module for 10k steps. Finally, in Stage 3, we use Wan2.2 5B~\cite{wan2025wan} as our base model and perform end-to-end training of the DiT model for 30k steps. Due to computational resource limitations, we process all training data into videos with 81 frames. In Stage 1 and Stage 2, all models were trained using AdamW with a learning rate of $10^{-4}$ . In Stage 3, the DiT model was trained with a learning rate of $10^{-5}$ , while the remaining components were 
trained with a learning rate of $5\times 10^{-5}$ .

\subsection{Evaluations and Comparisons}
We conduct a comparison with several  models, including Animate-X~\cite{tan2024animate}, AnimateAnyone~\cite{hu2024animate}, MimicMotion~\cite{zhang2024mimicmotion}, UniAnimate-DiT~\cite{wang2025unianimate},  Champ~\cite{zhu2024champ} and Wan-Animate~\cite{cheng2025wan}. We conduct experiments on cross reenactment and self reenactment to evaluate the performance of our model. For cross reenactment, we calculate several metrics on the constructed paired dataset, including PSNR, SSIM, LPIPS, FID and FVD .  For self reenactment, we assess the generation quality on the  first 50 entries of TikTok dataset  to examine the model’s performance in practical applications.
\begin{figure}[t!]
    \centering
    \includegraphics[width=0.45\textwidth]{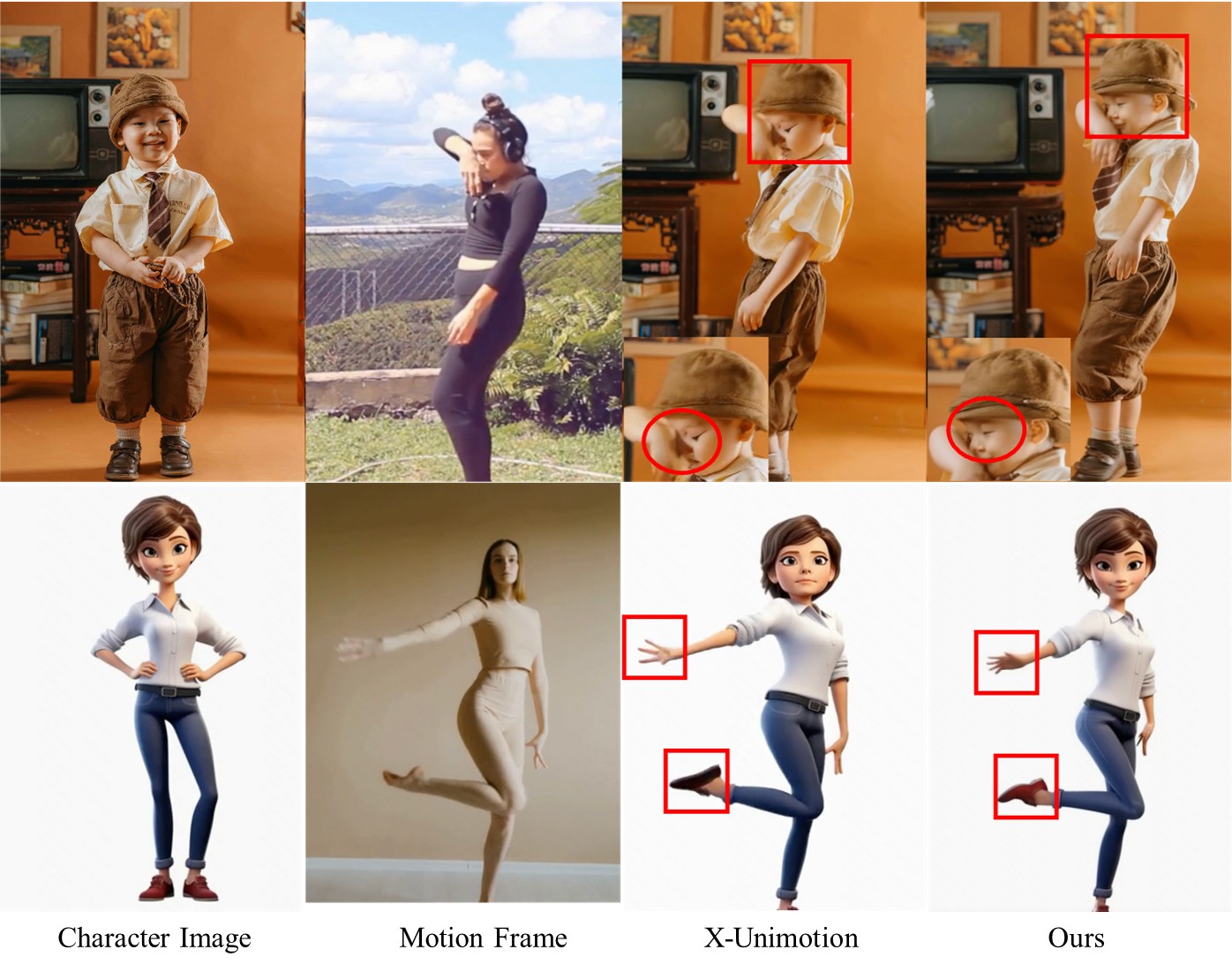} 
    \caption{Qualitative Results Compared with X-Unimotion. Since X-Unimotion has not released its source code, we conducted a visual comparison using the sample examples provided on the X-Unimotion homepage.}
    \label{fig:x-uni}
    \vspace{-1em}
\end{figure}
As shown in Table~\ref{tab:exp_1} and Figure~\ref{fig:exp1}, IM-Animation model demonstrates significant advantages in both cross reenactment and self reenactment tasks. In the cross reenactment task, IM-Animation achieves the highest scores in PSNR, LPIPS, and FID, indicating superior visual clarity and appeal. In SSIM and FVD, IM-Animation remains close in performance to Wan-Animate, which uses a larger base model and costs much more inference time. Since X-UniMotion has not yet been open-sourced, we provide a visual comparison of the cases available on its official website, as shown in Figure~\ref{fig:x-uni}. In the \yl{the} self-reenactment task, IM-Animation shows strong competitiveness in all major metrics. These results confirm that IM-Animation can maintain high performance and competitiveness. To demonstrate the robustness of IM-Animation, we present additional results on in-the-wild data in Figure~\ref{fig:in the wild performance}.

\subsection{Ablation Study}
We perform ablation experiments to systematically evaluate the contributions of different components of our model. This approach helps us understand the impact of each component on the overall performance and effectiveness of our method. We validate the effectiveness of several key modules in our model, including the 1D tokenizer-based motion encoder, which is compared to a direct convolutional module for encoding. In addition, we investigate the contributions of the mask token-based motion encoder, heatmap supervision, and the expression adapter.
\begin{table}[t!]\centering
    \caption{Quantitative evaluation of video generation on benchmark and TikTok  datasets. Values in bold represent the best performance, while underlined values indicate the second.
    }
    \label{tab:exp_1}

\resizebox{0.43\textwidth}{!}{
\large
\begin{tabular}{*{10}{c}}
    \toprule
    \multicolumn{6}{c}{\textbf{Cross Reenactment}} \\ 
    \midrule
    Methods & PSNR $\uparrow$ & SSIM $\uparrow$ & LPIPS $\downarrow$ & FID $\downarrow$ & FVD $\downarrow$ \\
    \midrule
    Champ~\cite{zhu2024champ} & 20.42 & 0.89 & 0.28 & 88.09 & 785.54 \\
    MimicMotion~\cite{zhang2024mimicmotion} & 20.01 & 0.88 & 0.30 & 97.86 & 901.55 \\
    AnimateAnyone~\cite{hu2024animate} & 19.24 & 0.86 & 0.31 & 85.24 & 690.15 \\
    UniAnimate-DiT~\cite{wang2025unianimate} & 21.97 & 0.90 & 0.25 & 71.61 & 436.80 \\
    Animate-X~\cite{tan2024animate} & 22.16 & 0.90 & 0.27 & 75.38 & 624.24 \\
    Wan-Animate~\cite{cheng2025wan} & \underline{22.72} & \textbf{0.92} & \underline{0.24} & \underline{59.10} & \textbf{267.71} \\
    \midrule
    IM-Animation (ours) & \textbf{22.87} & \underline{0.91} & \textbf{0.24} & \textbf{51.19} & \underline{270.42} \\ 
    \midrule\midrule
    \multicolumn{6}{c}{\textbf{Self Reenactment}} \\ 
    \midrule
    Methods & PSNR $\uparrow$ & SSIM $\uparrow$ & LPIPS $\downarrow$ & FID $\downarrow$ & FVD $\downarrow$ \\
    \midrule
    Champ~\cite{zhu2024champ} & 17.05 & 0.70 & 0.37 & 94.29 & 959.92 \\
    MimicMotion~\cite{zhang2024mimicmotion} & 16.81 & 0.71 & 0.39 & 92.04 & 927.46 \\
    AnimateAnyone~\cite{hu2024animate} & 18.13 & 0.71 & 0.34 & 89.37 & 793.61 \\
    UniAnimate-DiT~\cite{wang2025unianimate} & 19.76 & 0.75 & 0.29 & 71.32 & 504.19 \\
    Animate-X~\cite{tan2024animate} & 20.16 & 0.76 & 0.30 & 76.24 & 483.60 \\
    Wan-Animate~\cite{cheng2025wan} & \underline{21.12} & \textbf{0.78} &\underline {0.21} & \underline{36.78} & \underline{382.86} \\
    \midrule
    IM-Animation (ours) & \textbf{21.85} & \underline{0.76} & \textbf{0.20} & \textbf{35.98} & \textbf{351.70} \\ 
    \bottomrule
\end{tabular}

}

    \vspace{-1em}
\end{table}

As shown in the Figure~\ref{fig:abl} and Table~\ref{tab:abl1}, exp.A demonstrates that our motion encoder  extracts high-quality motion representations.
Compared to lightweight motion encoders, our model can effectively suppress the leakage of identity information and positional information of the original images. 
Exp.B shows that the expression adapter better preserves the facial expression details of characters. Exp.C indicates that our designed retargeting module can more effectively transfer motion sequences of characters with different postures to new characters. Finally, exp.D reveals that not adding extra middle supervision can lead to the generation of additional limbs or poor motion correspondences.
\begin{table}[t!]\centering
    \caption{Ablation study of different modules. We evaluated the effectiveness of the motion encoder design, retargeting module, and expression adapter through experiments .
    }
    \label{tab:abl1}
\resizebox{0.45\textwidth}{!}{
\footnotesize
\begin{tabular}{*{6}{c}}
\hline
Method          & PSNR* $\uparrow$ & SSIM $\uparrow$ & LPIPS $\downarrow$  & FVD $\downarrow$ & FID $\downarrow$ \\ \hline
{exp A.} \emph{w/o}  1d tokenizer-based motion encoder
  & 19.67 & 0.84 & 0.29 & 81.51 & 543.66 \\
{exp B.} \emph{w/o} expression adapter  & 22.01 & 0.89 & 0.26 & 69.03 & 374.90 \\
{exp C.} \emph{w/o} mask token-based retargeting module
 & {21.03} & {0.88} & {0.27} & {78.32} & {561.89} \\
 {exp D.} \emph{w/o} middle supervision
 & {21.54} & {0.89} & {0.25} & {72.80} & {443.37} \\
 {exp E.} full model
 & \textbf{22.87} & \textbf{0.91} & \textbf{0.24} & \textbf{51.19} & \textbf{270.42} \\
 \hline
\end{tabular}
}

    \vspace{-1em}
\end{table}

\begin{table}[t!]\centering
    \caption{Ablation study of different training strategy. We evaluated the performance of different granularity levels of training division strategies alongside the strategy we ultimately adopted.
    }
    \label{tab:abl2}
\resizebox{0.46\textwidth}{!}{
\footnotesize
\begin{tabular}{*{6}{c}}
\hline
Method          & PSNR* $\uparrow$ & SSIM $\uparrow$ & LPIPS $\downarrow$  & FVD $\downarrow$ & FID $\downarrow$ \\ \hline
End-to-End training
  & 19.40 & 0.73 & 0.37 & 78.89 & 566.04 \\

 2 Stages training
 & {20.65} & {0.79} & {0.30} & {67.11} & {334.96} \\
 3 Stages training
  & \textbf{22.97} & \textbf{0.91} & \textbf{0.24} & \textbf{51.19} & \textbf{270.42}\\
 \hline
\end{tabular}
}

    \vspace{-1em}
\end{table}
In addition, we evaluate the effectiveness of our three-stage training approach. This evaluation involves comparing the current three-stage training setup with a joint training method that retains Stage 2 and Stage 3, as well as an end-to-end training approach. All comparisons are conducted using the DiT training framework for 30k steps.

As shown in Table~\ref{tab:abl2}, our model can quickly converge on the metrics of the three-stage training with a shorter number of DiT training steps. We believe that the motion representation and retargeting module, after initial training, can provide better guidance to the DiT compared to randomly initialized modules. Additionally, during the end-to-end training process, the flow matching decoder can offer improved supervision for the motion representation module.

\subsection{User Study}
We \yl{conducted} a user study to assess participants' subjective preferences regarding the generated videos. We selected 8 benchmark videos from the cross-ID task and invited 20 participants to evaluate them. Participants rate the videos based on several criteria: motion and expression quality (MQ), identity preservation (IP), video reality (VA), and overall video quality (OA).
\begin{table}[t!]\centering
    \caption{User study results.
    }
    \label{tab:user_study}

\renewcommand{\arraystretch}{0.9} 
\resizebox{0.45\textwidth}{!}{
\tiny 
\begin{tabular}{*{5}{c}}
    \hline
    \multicolumn{5}{c}{\textbf{User Study}} \\ 
    \hline
    Methods & MQ $\uparrow$ & IP $\uparrow$ & VR $\uparrow$ & OA $\uparrow$\\
    \hline
    Animate-X & 3.95 & 3.35 & 3.25 & 3.60 \\
    MimicMotion & 3.20 & 3.35 & 3.20 & 3.30  \\
    AnimateAnyone & 2.55 & 3.15 & 2.9 & 2.70  \\
    UniAnimate-DiT & 3.90 & 3.85 & 3.90 & 3.75  \\
    Champ & 2.75 & 2.80 & 2.05 & 2.10  \\
    Wan-Animate & 4.35 & 4.40 & 4.35 & 4.35  \\
    \hline
    IM-Animation (ours) & \textbf{4.55} & \textbf{4.45} & \textbf{4.35} & \textbf{4.40}  \\ 
    \hline
\end{tabular}
}
\vspace{-1.em}

\vspace{1.5mm}
\end{table}
Each participant watched several video clips, one of them generated by our method and others by baseline methods.This allowed participants to evaluate multiple pairs of videos, leading to a comprehensive comparison.

The results in Table~\ref{tab:user_study} indicate a strong preference for our method among participants. Our experiments are consistent with both quantitative and qualitative evaluations, further demonstrating the effectiveness of our approach in meeting user expectations for high-quality human video generation.
\section{Conclusion}

{We introduce IM-Animation, a diffusion-based character animation framework built upon a compact implicit motion representation and a mask token based retargeting module. By encoding per-frame dynamics into 1D motion tokens, our method mitigates identity leakage from the driving video. The mask-token bottleneck further separates identity information from the source image, leading to coherent pose retargeting even under significant mismatches in body shape, scale, and spatial layout. Extensive experiments demonstrate that IM-Animation achieves competitive or superior motion fidelity, identity preservation, and visual quality compared with existing 
approaches.}


However, we acknowledge that our current strategy does not fully guarantee that facial expression features are entirely independent of the original identity. {Designing specific facial retargeting module may be a potential solution.} Additionally, allowing users to control the camera perspective is an exciting direction for future work. {Integrating our implicit motion control with explicit 3D scene point cloud representation possibly solve this issue.}

{
{
    \small
    \bibliographystyle{ieeenat_fullname}
    \bibliography{main}
}
\clearpage
\setcounter{page}{1}
\maketitlesupplementary

\section{Implementation Details}
\label{sec:Implementation Details}
In the following section, we outline the training strategies for IM-Animation, designed to achieve efficient convergence with limited resources. Unlike existing implicit video-driven methods, our approach consists of three stages. In Stage 1, we train the motion encoder, fine-tuning the pretrained encoder and quantizer from TiTok~\cite{yu2024an} using self-reconstructed joint maps for supervision. Stage 2 involves joint training of the motion encoder and the retargeting module, incorporating data augmentation techniques and constructing action-consistent but identity-inconsistent paired datasets. Finally, in Stage 3, we inject motion and expression control signals into the DiT model for end-to-end training, while maintaining supervision from the joint decoder. This structured approach aims to enhance the model's performance in complex scenarios and ensure effective learning of motion representations.

\textbf{Stage 1: Motion encoder training stage.}
As mentioned in the main text, our subsequent experiments are based on the pretrained checkpoints provided by TiTok.

TiTok's original design is based on image reconstruction and generation tasks, and the features extracted by its encoder tend to capture more detailed image semantics than desired. During the motion representation phase, we aim to minimize identity detail leakage to the downstream DiT model. To achieve this, we have meticulously designed a decoder that transforms the compressed tokens, of which we mentioned there are 32 in the main text, into a joint map.
\begin{figure*}[t!]
    \centering
    \includegraphics[width=1\textwidth]{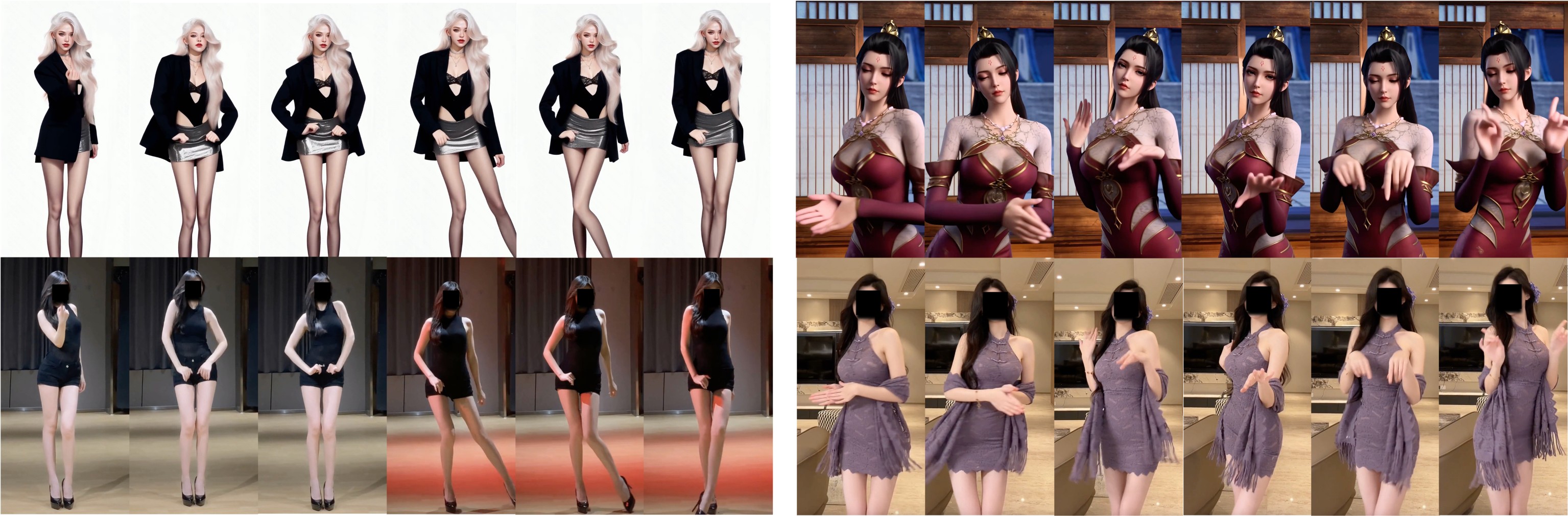} 
    \caption{Samples of synthesized data . The first row is the reference video and the second row is the target video.}
    \label{fig:supp_1}
    \vspace{-1em}
\end{figure*}
\begin{figure*}[t!]
    \centering
    \includegraphics[width=1\textwidth]{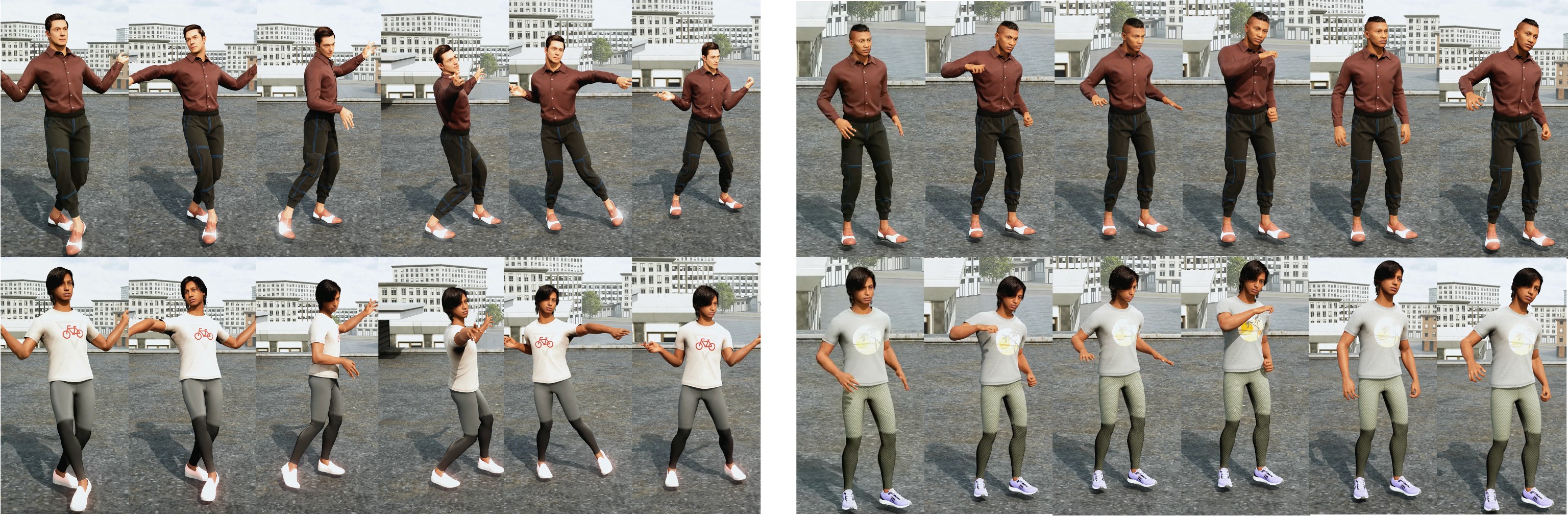} 
    \caption{Samples of UE data . The first row is the reference video and the second row is the target video.}
    \label{fig:supp_2}
    \vspace{-1em}
\end{figure*}
\begin{figure}[t!]
    \centering
    \includegraphics[width=0.45\textwidth]{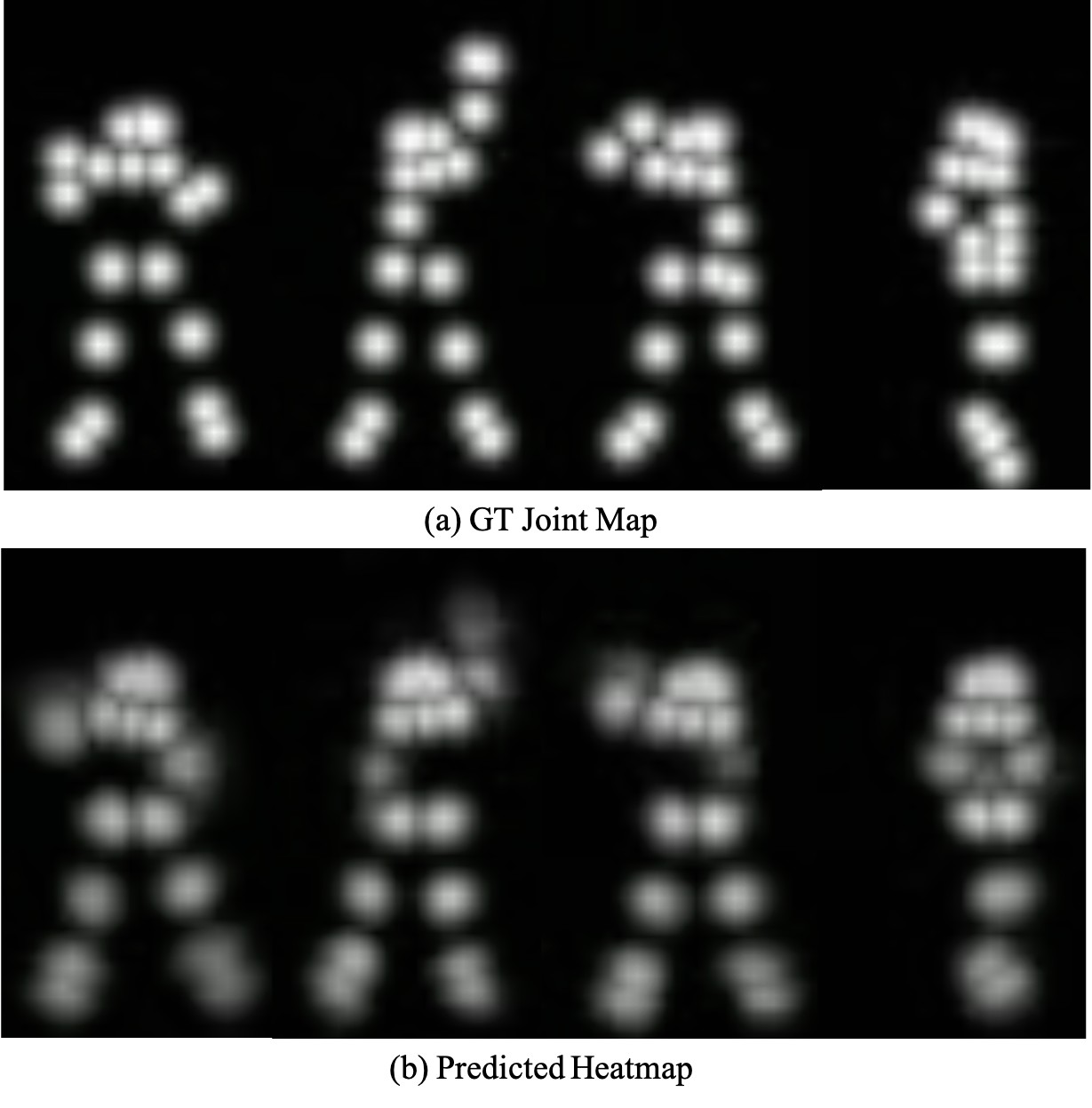} 
    \caption{Visualization of Joint Heatmap.}
    \label{fig:supp_3}
    \vspace{-1em}
\end{figure}
In this process, we restructure the motion decoder. We align the number of mask tokens with the quantity of joint supervision. Specifically, we select the 20 body joints from DWPose~\cite{yang2023effective} along with all hand joints as supervision points. After obtaining the corresponding joint tokens, we employ a series of convolutional layers to upsample the mask tokens to the scale of the ground truth joint map.
\begin{equation}
Loss_{motion} = \frac{1}{T} \sum_{t=1}^{T} \left( H_{motion,t} - H_{\text{gt},t} \right)^2
\end{equation}
Here, $T$ represents the total number of frames in the video sequence, $H_{motion,t}$ is the heatmap generated by the model for the $t$-th frame, and $H_{\text{gt},t}$ is the corresponding ground truth heatmap. At this stage, we do not design data augmentation or similar self-supervised retargeting training constructs like X-UniMotion~\cite{song2025x}. In other words, this stage is more akin to training a keypoint prediction process.

\textbf{Stage 2: Retargeting training stage.}
In the implementation of the retargeting module, we first compress the source image into latent space using the VAE from Wan2.2~\cite{wan2025wan} before performing patchification. In our design, the number of mask tokens is aligned with the number of patches from the source image. This approach allows us to overcome the limitations of the patch grid while facilitating channel concatenation during the subsequent process of controlling condition injection.

In this phase, unlike the first stage where we directly regress the heatmap, we employ random data augmentation to enable the model to learn retargeting across different IDs and at larger scales. After applying random color transformations, we randomly crop or scale the original video, and then supervise the model using the original video. The loss function of this stage is defined as follow,
\begin{equation}
    L_{\text{retarget}} = \frac{1}{T} \sum_{t=1}^{T} (H_{\text{retarget,t}} - H_{\text{gt,t}})^2
\end{equation}
where $T$ denotes the total number of time steps, $H_{\text{retarget,t}}$ is the generated retargeting heatmap, and $H_{\text{gt,t}}$ is the ground truth heatmap at time step $t$. By averaging the squared differences between the predicted heatmap and the corresponding ground truth heatmap at each time step, $L_{\text{retarget}}$ aims to minimize the disparity between the generated and true heatmaps, thereby improving the model's performance in the retargeting task.

\textbf{Stage 3: End-to-End training stage.}
In the final stage, we conduct end-to-end training. In this part, we use additional synthetic data to enhance the data scale. Our method for injecting control signals has been thoroughly described in the main text. To avoid disrupting the pre-trained capabilities of DiT, we reuse the weights from the original checkpoint for the embedding part corresponding to the channel-wise concatenation, while initializing the region part to zero. The original Wan2.2 5B model contains 30 DiT blocks. We insert an expression block after every 6 DiT blocks, employing a skip connection technique to prevent disrupting the original capabilities of the DiT.
At this stage, we also employ intermediate supervision for training. For the diffusion loss function,
\begin{equation}
    L_{DiT}={\rm{E}}_{t,x_0^{1:N},y,\epsilon} \left \| \epsilon - \epsilon_{\theta} (\sqrt{\bar{\alpha} _{t} } x_0^{1:N} + \sqrt{1 - \bar{\alpha}_{t} } \epsilon ,t, y) \right \| ^ 2,
\end{equation}
where \( t \) is sampled from the range \( [1, T] \) (denoting the denoising steps), \( \epsilon \) represents the random noise, \( y \) denotes the text prompts, and \( x_0^{1:N} \) refers to the video data comprising \( N \) frames. 

We also employ the aforementioned intermediate supervision to accelerate convergence. The loss function for the entire Stage Three is defined as
\begin{equation}
    L_{total} = L_{DiT} + \alpha\cdot L_{retarget}
\end{equation}
During our training process, \( \alpha \) is set to 10.

\section{Dataset Details}
We utilize Kling~\cite{kwai2025kling} to synthesize data that maintains consistent motion across diverse identities. Throughout the synthesis process, we engineered data featuring a range of spatial characteristics and body types, followed by a meticulous manual selection of the outputs. Sample outputs of the generated data are presented in Figure ~\ref{fig:supp_1}.

For the UE data, similar to the approach taken in ReCameraMaster~\cite{bai2025recammaster}, we synthesized data from various camera positions and an array of scenes. This data synthesis strategy enhances our model's ability to adapt to different viewpoints and environmental variations, thereby improving both the quality and diversity of the generated character animations.

\section{Visualization of Joint Map}
To validate the effectiveness of our intermediate supervision, we present visual results of the joint decoder outputs following the redirection process, as illustrated in Figure \ref{fig:supp_3}. The heatmaps generated by our decoder demonstrate a strong correspondence with the ground truth, indicating a high level of accuracy in the predictions. This alignment not only underscores the robustness of our model, but also highlights the efficacy of the intermediate supervision strategy in enhancing the learning process.

The visual comparison reveals that the decoder successfully captures the intricate details of the target output, suggesting that our approach effectively mitigates discrepancies that typically arise during the generation process. Such results provide compelling evidence that our method contributes to improved performance in generating high-fidelity character animations.

\section{More Ablation Study}
\begin{table}[t!]\centering
    \caption{Motion Encoder Comparsion .
    }
    \label{tab:abl3}
\resizebox{0.45\textwidth}{!}{
\footnotesize
\begin{tabular}{*{6}{c}}
\hline
Method          & PSNR* $\uparrow$ & SSIM $\uparrow$ & LPIPS $\downarrow$  & FVD $\downarrow$ & FID $\downarrow$ \\ \hline
Wan + VAE motion encoder
  & 17.14 & 0.76 & 0.37 & 94.44 & 653.30 \\
  full model
 & \textbf{22.87} & \textbf{0.91} & \textbf{0.24} & \textbf{51.19} & \textbf{270.42} \\
 \hline
\end{tabular}
}

\end{table}
\begin{table}[t!]\centering
    \caption{Retargeting Module Comparsion .
    }
    \label{tab:abl4}
\resizebox{0.45\textwidth}{!}{
\footnotesize
\begin{tabular}{*{6}{c}}
\hline
Method          & PSNR* $\uparrow$ & SSIM $\uparrow$ & LPIPS $\downarrow$  & FVD $\downarrow$ & FID $\downarrow$ \\ \hline
Wan + SA retargeting module
  & 19.89 & 0.82 & 0.26 & 68.99 & 477.63 \\
  full model
 & \textbf{22.87} & \textbf{0.91} & \textbf{0.24} & \textbf{51.19} & \textbf{270.42} \\
 \hline
\end{tabular}
}

\end{table}
\begin{figure}[t!]
    \centering
    \includegraphics[width=0.45\textwidth]{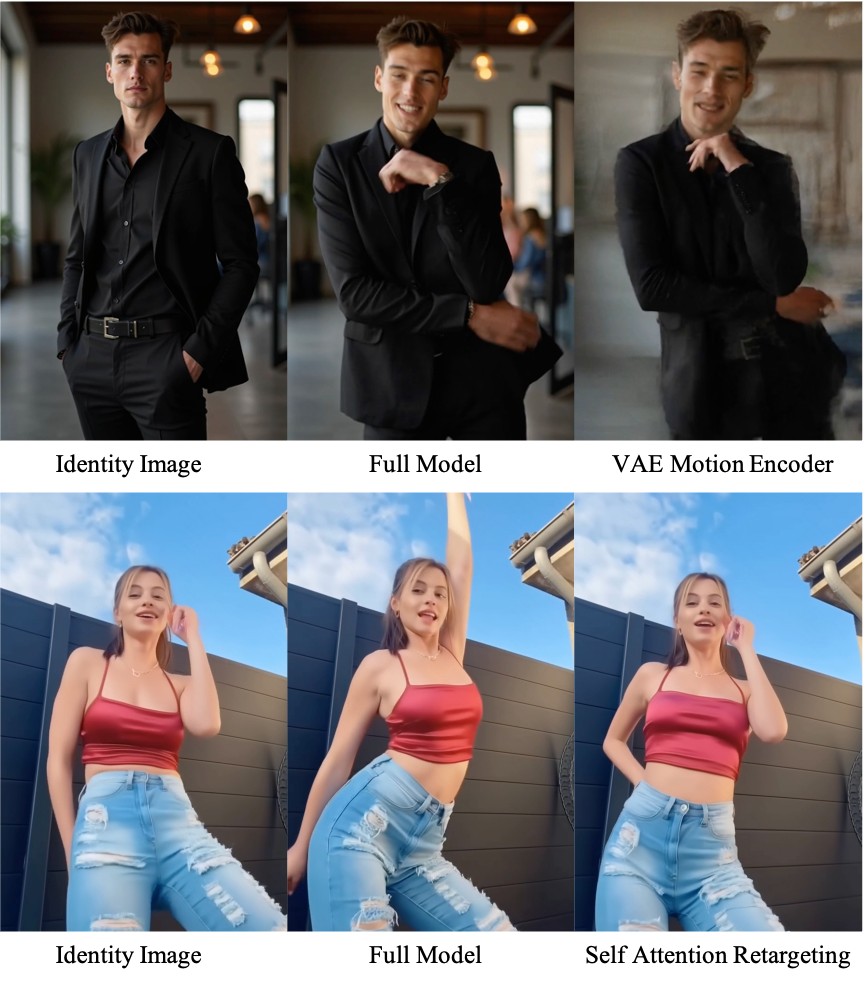} 
    \caption{Ablation Study.}
    \label{fig:supp_4}
\end{figure}
For the motion encoder, we conduct more fine-grained ablation experiments. We compare our motion encoder with the results of directly inputting latent variables compressed by VAE into the downstream model for retargeting. This comparison allows us to more clearly assess the advantages of the motion encoder in terms of generation quality and performance.

As shown in the Figure \ref{fig:supp_4} and Table \ref{tab:abl3}, directly using VAE for encoding results in background information being directly encoded into the motion tokens, leading to a decline in generation quality. In contrast, our method effectively prevents the leakage of semantic information from the motion tokens.

In addition, we conduct a fine-grained validation of the effectiveness of the mask token-based retargeting module. In fact, we experiment with the retargeting implementation proposed by X-UniMotion, which utilizes self-attention to combine two sources of tokens. Although this implementation is not open-sourced, we develop a similar version and conduct comparative experiments. As shown in the table, we find that in some cases, this type of retargeting leads to an increased probability of implicit representation control failure, resulting in generated videos that tend to maintain static motions which can be found in Figure \ref{fig:supp_4}.
\begin{figure*}[t!]
    \centering
    \includegraphics[width=0.95\textwidth]{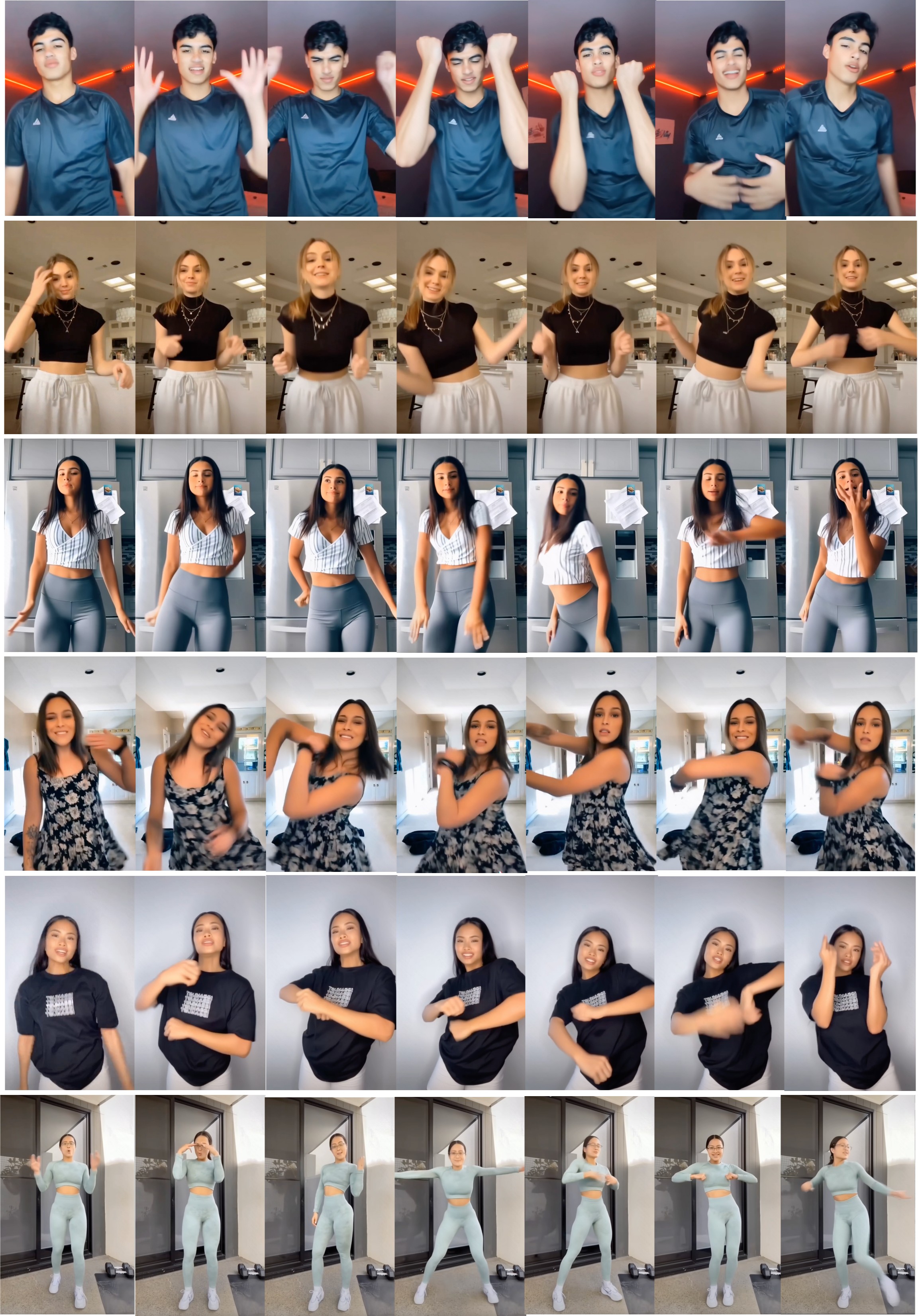} 
    \caption{More visualization results on TikTok dataset.}
    \label{fig:supp_5}
    \vspace{-1em}
\end{figure*}
\begin{figure*}[t!]
    \centering
    \includegraphics[width=0.95\textwidth]{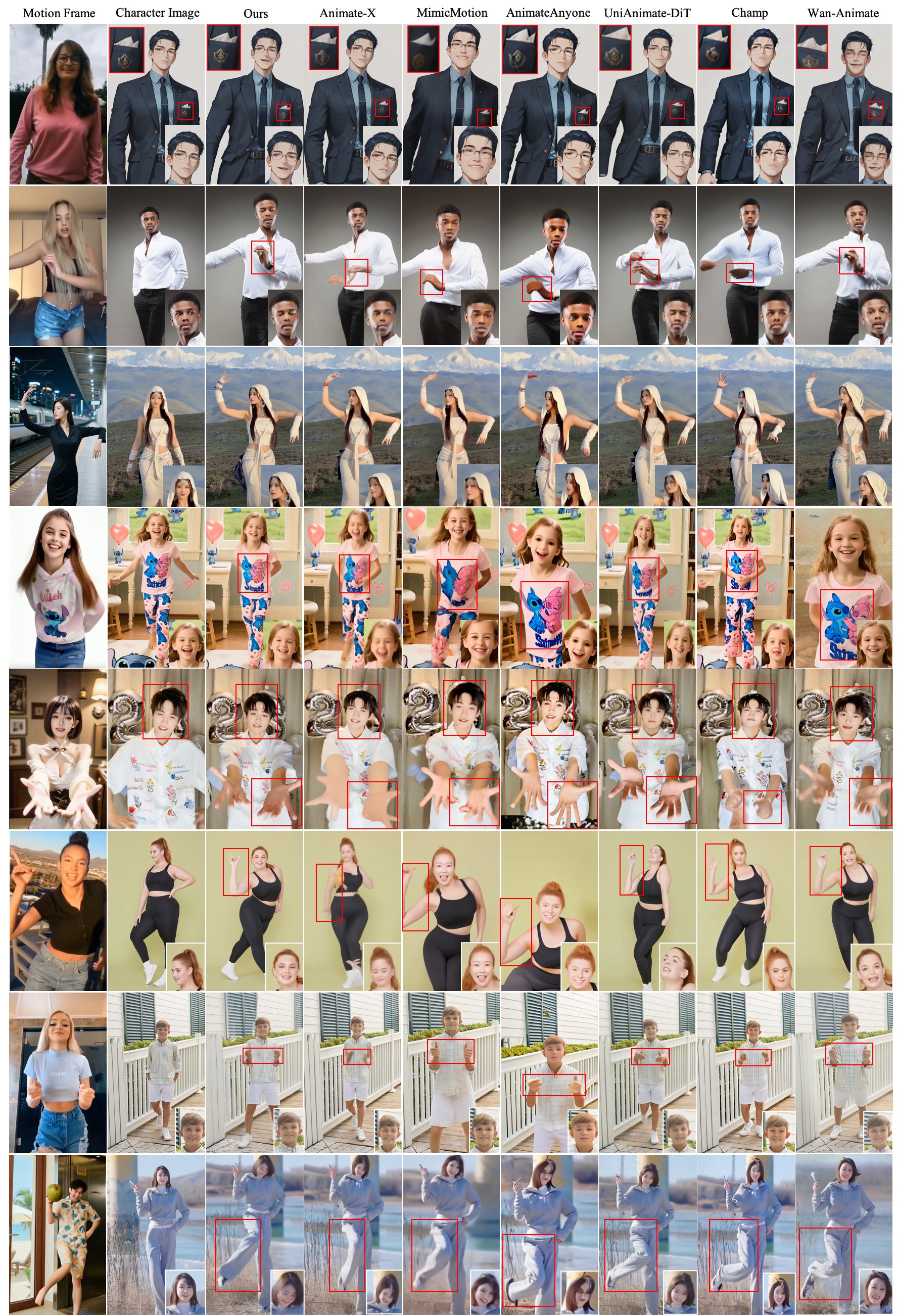} 
    \caption{More visualization results of Comparsion Results.}
    \label{fig:supp_6}
    \vspace{-1em}
\end{figure*}
\begin{figure*}[t!]
    \centering
    \includegraphics[width=0.95\textwidth]{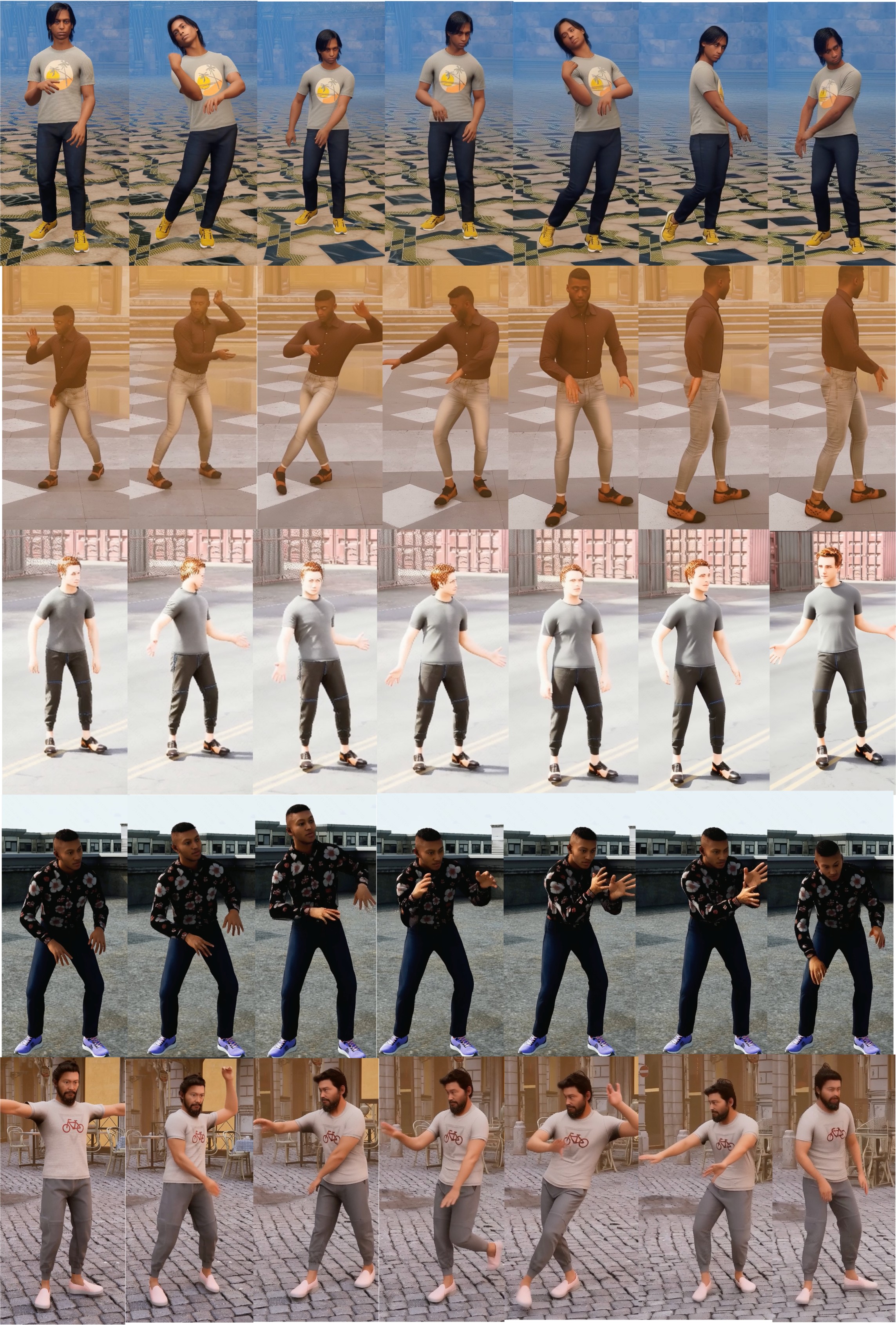} 
    \caption{More visualization results on UE dataset.}
    \label{fig:supp_8}
    \vspace{-1em}
\end{figure*}
\begin{figure*}[t!]
    \centering
    \includegraphics[width=0.95\textwidth]{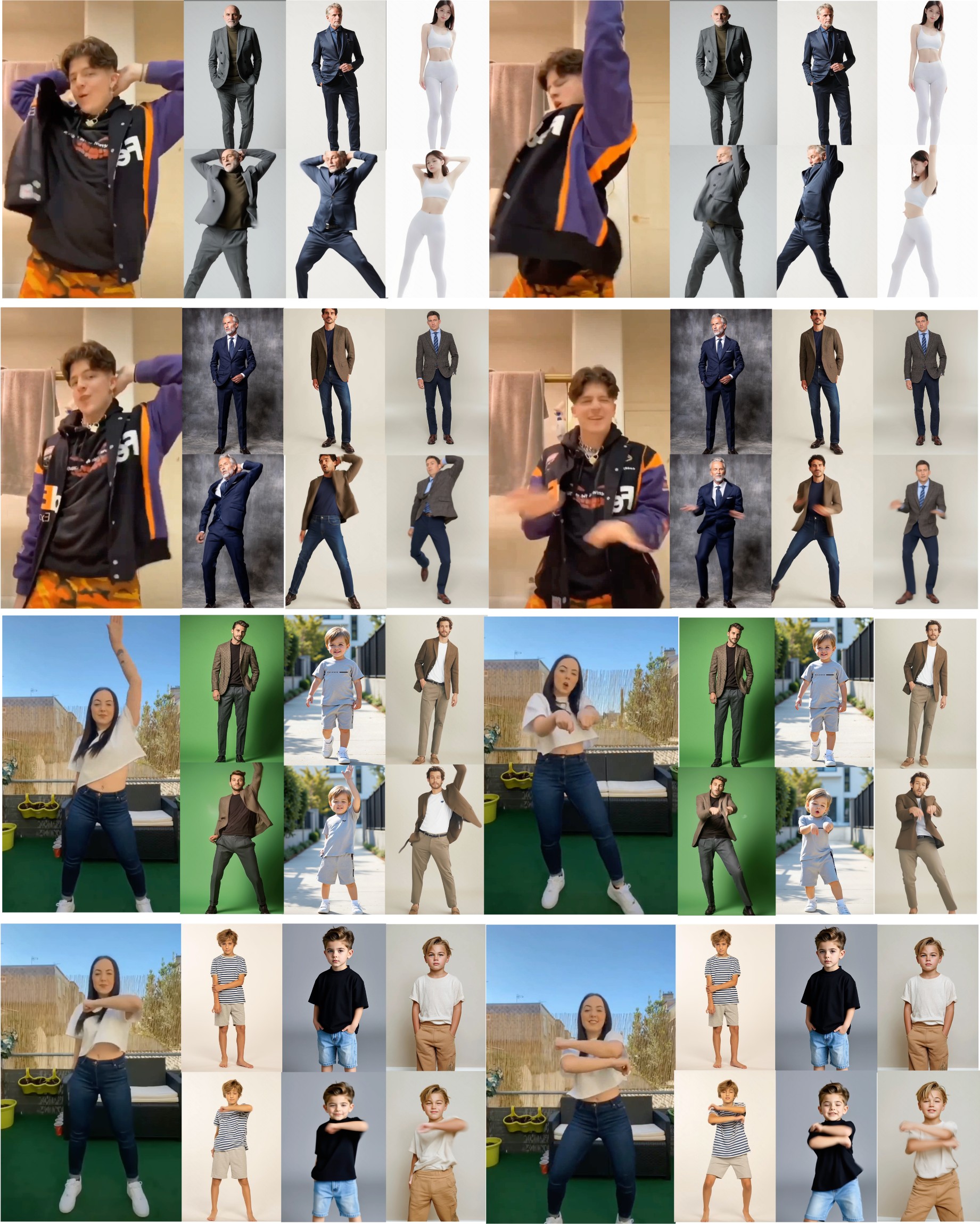} 
    \caption{More visualization results on in the wild data}
    \label{fig:supp_9}
    \vspace{-1em}
\end{figure*}
\section{More Visualization results.}
Here, as shown in the Figure \ref{fig:supp_5}, we provide some generated results of IM-Animation on the TikTok dataset.

Additionally, as shown in Figure \ref{fig:supp_6}, we present more qualitative results here compared to Animate-X~\cite{tan2024animate}, MimicMotion~\cite{zhang2024mimicmotion}, AnimateAnyone~\cite{hu2024animate}, UniAnimate-DiT~\cite{wang2025unianimate}, Champ~\cite{zhu2024champ} and Wan-Animate~\cite{cheng2025wan}, highlighting that our model better preserves certain details compared to those in the main text. We also demonstrate that our model performs well in cases where explicit driven retargeting fails. In Figure \ref{fig:supp_8}, we also present the generation quality on the synthetic UE dataset.
In the Figure~\ref{fig:supp_9}, we also present additional visualization results.


\end{document}